\documentclass[10pt,twocolumn,letterpaper]{article}

\usepackage[pagenumbers]{cvpr}

\usepackage{cuted} %

\usepackage{currfile} %

\usepackage{caption} %

\usepackage{bm}
\usepackage{makecell}
\newcommand{\ours}{CoRelate\xspace}
\newcommand{\Rset}{\mathcal{R}}
\newcommand{\Mset}{\mathcal{M}}
\DeclareMathOperator{\FK}{FK}

\usepackage{placeins} %

\definecolor{cvprblue}{rgb}{0.21,0.49,0.74}
\usepackage[pagebackref,breaklinks,colorlinks,allcolors=cvprblue]{hyperref}

\def\confName{CVPR}
\def\confYear{2027}

\title{CoRelate: Coordinated Human--Object Motion Synthesis\\ through Interaction Relations}

\author{
Zhiwen Qiu\textsuperscript{1,*} \quad
Daisy Lin\textsuperscript{1,*} \quad
Diyu Zhou\textsuperscript{1} \\
Jialin Wu\textsuperscript{1} \quad
Rosita Ji\textsuperscript{1} \quad
Saleh Kalantari\textsuperscript{1} \\
\textsuperscript{1}Cornell University \\
{\small \textsuperscript{*}Equal contribution}
}

\begin{document}
\makeatletter\g@addto@macro\@maketitle{\vspace*{-11pt}}\makeatother
\maketitle
\begin{strip}
\centering
\includegraphics[width=\linewidth]{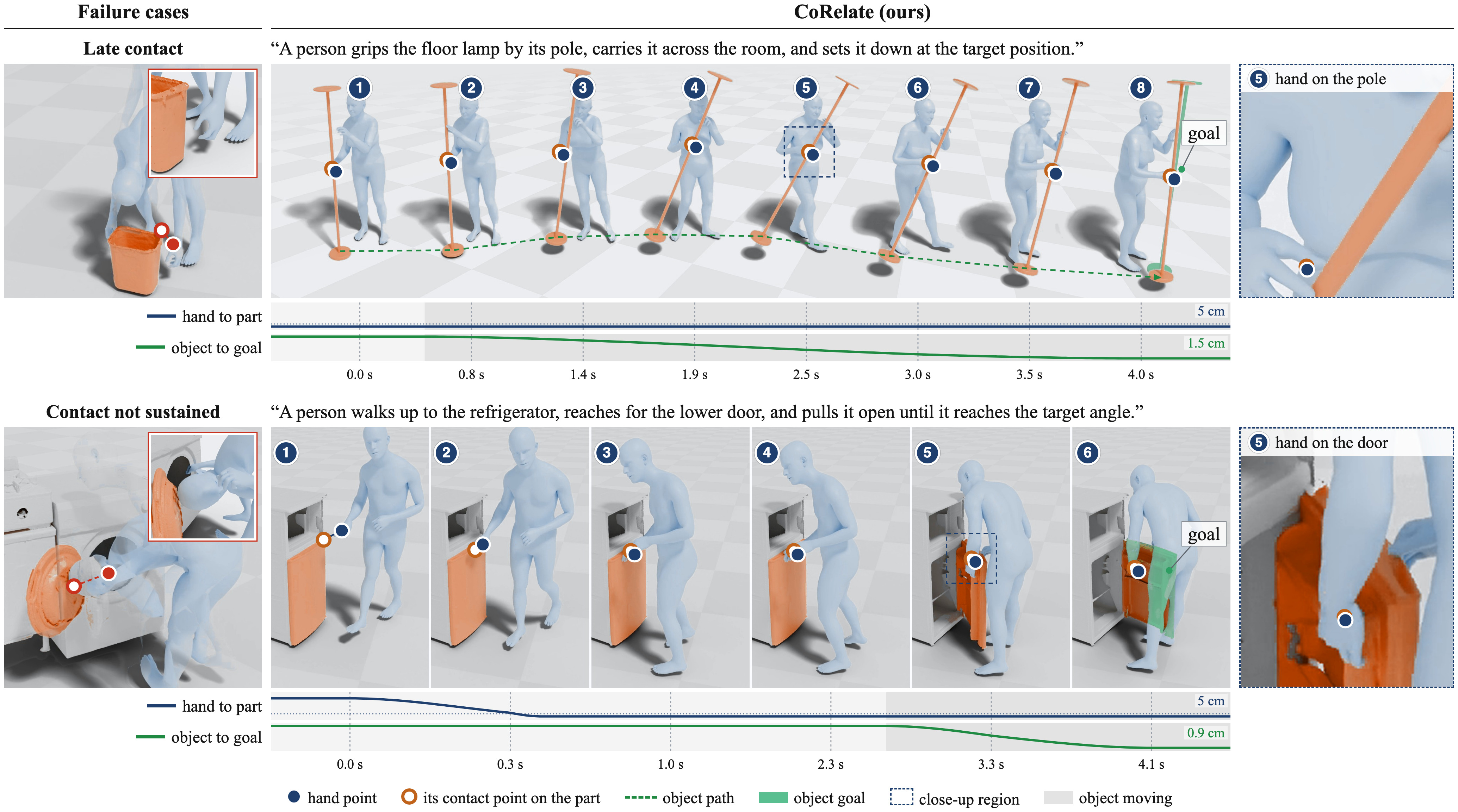}
\fontsize{8}{8.4}\selectfont
\captionsetup{hypcap=false,skip=2pt,justification=raggedright,singlelinecheck=false}
\captionof{figure}{\textbf{Coordinated human--object motion generation.}
Reaching an object goal does not ensure timely or sustained hand contact, as the object may begin moving before contact is established or continue moving after contact is lost (left). \ours jointly refines human and object motion using hand--part relation feedback and object-goal task feedback to maintain contact during object motion while meeting the specified goal (right).}
\captionsetup{hypcap=true}
\label{fig:teaser}%
\end{strip}

\begin{abstract}
Human--object motion synthesis requires coordinated human participation as objects move toward specified goals. However, reaching object targets does not ensure that contact is established and maintained, while adjustments that improve contact may interfere with task completion. We introduce CoRelate, a diffusion framework that coordinates full-body human--object motion through explicit interaction relations and task feedback for rigid and articulated objects. We represent intended participation through hand--part relations that specify the participating hand, its part-relative reference position, and participation interval. These relations serve both as generation conditions and as references for measuring interaction realization. At each denoising step, a temporal refiner trained jointly with the generator combines relation deviations, task residuals, and motion features from the current prediction to jointly update human and object motion across the sequence. After generation, temporal motion correction addresses remaining contact failures by optimizing local motion segments while preserving object trajectories and protected human motion. Experiments demonstrate improved object-goal satisfaction, contact F1, and complete interaction success over the evaluated baselines. Ablation studies support the benefits of relation conditioning and the complementary roles of task and relation feedback.
\end{abstract}

\section{Introduction}
\label{sec:intro}

Generating full-body human--object interaction is central to character animation, interactive simulation, and behavior modeling for embodied agents and robotic systems \cite{li2023omomo,li2024chois,xu2025intermimic}. Purposeful interaction requires the human and object to move coherently over time while accomplishing a requested task goal. Sparse object waypoints offer a practical way to specify such tasks without prescribing a complete motion sequence, by constraining selected object states while leaving substantial freedom in how the person participates and how the object moves between waypoints \cite{li2024chois,wang2026mamihoi}. Successful human--object motion synthesis must therefore realize both what the object accomplishes and how the human participates in the same generated motion (\cref{fig:teaser}).

Prior work on joint human--object generation infers both motions from language or motion conditions \cite{peng2025hoidiff,diller2024cghoi}, with additional control through initial states and sparse object waypoints \cite{li2024chois,wang2026mamihoi}. Contact and relation modeling can improve human--object coordination \cite{xue2025rog,li2026pami}. Geometry-aware guidance and learned refinement further improve geometric alignment and motion coherence \cite{wang2026mamihoi,li2026pami,pi2025coda}. However, in operations requiring hand contact, an object may begin moving before contact is established or continue moving while the hand remains separated from its surface or loses contact. Such failures can persist even when the supplied object targets are satisfied. Reducing relation error alone does not determine an appropriate correction under sparse task constraints, since the discrepancy can be reduced by adjusting human motion, intermediate object motion, or both. Task residuals provide an explicit signal for coordinating these adjustments with the supplied object targets, as moving an object toward the hand may undo an already reached target or help correct an overshoot. Moreover, an update that improves contact in one interval may disrupt participation or motion continuity elsewhere. We use representations of intended participation as a reference for measuring its realization and guiding motion updates together with task feedback. This leads to our research question: How can we represent and use interaction relations to coordinate human--object motion that establishes and sustains contact while satisfying object goals and preserving motion quality?

We present \ours, a diffusion framework that coordinates human--object motion under sparse object goals using explicit interaction relations and task feedback. From an initial motion proposal and task geometry, we construct hand--part relations specifying which hand participates, its reference position relative to the object part, and its participation interval. These relations serve both as generation conditions and as references for measuring interaction realization. During denoising, relation deviations measure departures from the selected hand--part arrangement, while task residuals measure errors at the supplied object targets. A temporal refiner trained jointly with the generator, combines these signals with motion features to predict coordinated human and object updates across the sequence. Relations and sparse waypoints remain fixed, while feedback is recomputed from the evolving human and object trajectories at each denoising step. A common part-relative interface supports rigid and articulated interactions, with object updates applied to rigid poses and articulated joint states, respectively. After joint generation and refinement, segment-level temporal motion correction addresses remaining contact failures. At this stage, we fix the generated object trajectory and a continuous solver adjusts local hand motion and preserves protected human motion. By distributing corrections across neighboring frames, the solver improves contact while limiting abrupt wrist changes. Our contributions are:
\begin{itemize}
  \item A relation representation that specifies where and when each hand participates to condition generation and evaluate interaction realization for rigid and articulated objects.
  \item A feedback-driven refiner that integrates relation deviations, task residuals, and motion features from current predictions to jointly update human and object motion.
  \item A temporal motion correction method that reduces residual contact failures while preserving object trajectories and human motion.
\end{itemize}

\section{Related Work}
\label{sec:relatedwork}

\noindent\textbf{Full-body human--object motion generation.}
Human--object generation builds on text-conditioned motion synthesis \cite{guo2022humanml3d,tevet2023mdm,chen2023mld,zhang2023t2mgpt,jiang2023motiongpt,lu2023humantomato} and controllable generation \cite{karunratanakul2023gmd,xie2024omnicontrol,zhang2023controlnet}. For moving objects, human motion can be generated from complete object trajectories \cite{li2023omomo}, while task-conditioned methods synthesize interactions from higher-level specifications \cite{ghosh2022imos,li2024toho}. Text-driven joint generators infer both human and object motion through coupled motion streams, contact guidance, and relation intervention \cite{peng2025hoidiff,diller2024cghoi,wu2024thor,song2024hoianimator}. For spatial control, CHOIS \cite{li2024chois} jointly generates human and object motion from language, initial states, and object waypoints. Related approaches introduce geometry-aware spatial control \cite{wang2026mamihoi} and flexible conditions over SE(3) trajectories \cite{lei2026wmm}. For articulated manipulation, CoDA \cite{pi2025coda} coordinates separate body and hand motion priors. This coordination allows whole-body posture to adapt to hand-level interaction constraints on movable object parts. Other articulated generators combine locomotion and manipulation \cite{zhang2026giraf} or use surface keypoint trajectories to represent multiple objects and articulation \cite{peng2026skhoi}.

\noindent\textbf{Interaction relations and dynamics.}
Geometric representations based on sampled points, interaction fields, and surface anchors provide spatial context for human--object interaction \cite{prokudin2019bps,kulkarni2024nifty,li2026contahoi}. Relative geometry and contact representations can support joint motion reconstruction \cite{xie2022chore,nam2024contho,huo2024stackflow}, whole-body grasping and dexterous manipulation \cite{taheri2022goal,wu2022saga,zhang2021manipnet,zhang2024artigrasp}, and text-driven hand motion \cite{cha2024text2hoi,christen2024diffh2o,li2025latenthoi,huang2025hoigpt}. Contact prediction \cite{peng2025hoidiff,diller2024cghoi,ma2024contactaware}, affordance models \cite{li2025hoipage,su2026uniffordinteractor}, and temporal contact models \cite{adossehoun2026when,song2026jointhoi} provide useful cues about where and when interaction should occur. Joint-level relation models \cite{zeng2025chainhoi} and part-relative motion representations with learned voting \cite{li2026pami} can encode human--object coupling within generation. For example, ROG \cite{xue2025rog} learns a spatiotemporal prior over interactive distance fields, where it refines the field computed from the current motion prediction and uses the refined relations to guide motion updates during denoising. SK-HOI \cite{peng2026skhoi} predicts a spatiotemporal contact distance field from generated object trajectories and uses it to guide human-motion synthesis through noise optimization. HOI-Dyn \cite{wu2025hoidyn} learns object responses to human motion and uses response consistency to supervise joint generation. Our method uses the selected local hand--part relations both to condition generation and to provide a reference for measuring participation. The realization is evaluated against the evolving motion prediction and the resulting deviations are combined with task residuals to enhance learned human--object updates within denoising.

\noindent\textbf{Geometric guidance and motion refinement.}
Geometric constraints are incorporated through contact supervision \cite{li2024chois,peng2025hoidiff,diller2024cghoi}, sampling guidance \cite{karunratanakul2023gmd,xue2025rog,song2026jointhoi}, and diffusion-noise optimization \cite{karunratanakul2024dno,ron2026hoidini,pi2025coda,peng2026skhoi} to improve contact alignment and spatial accuracy. Learned hand-motion models typically use local object surfaces and temporal hand--object correspondences to synthesize or refine poses that reduce penetration and contact inconsistencies \cite{taheri2024grip,zhou2022toch,zhou2024gears}. For full-body interaction, contact-relative motion prediction can correct denoised human--object sequences within diffusion \cite{xu2023interdiff}. For instance, MaMi-HOI \cite{wang2026mamihoi} combines geometry-aware feature adaptation with a spatially conditioned whole-body adapter to reduce the motion stiffness that local contact refinement can introduce. PAMI \cite{li2026pami} recursively updates human and object motion after initial coarse generation using surface geometry and temporal attention, coupled with body support, trajectory smoothness, and preservation losses to discourage unwanted motion changes. CoRelate uses hand--part references and recomputes their realization errors together with sparse-target residuals at each denoising step. A learned refiner uses both signals to jointly update human motion and object states. For remaining local failures after generation, we solve continuous arm segments while retaining object motion and restricting edits to the selected human joints and time intervals.

\section{Method}
\label{sec:method}

\begin{figure*}[t]
  \centering
  \includegraphics[width=\linewidth]{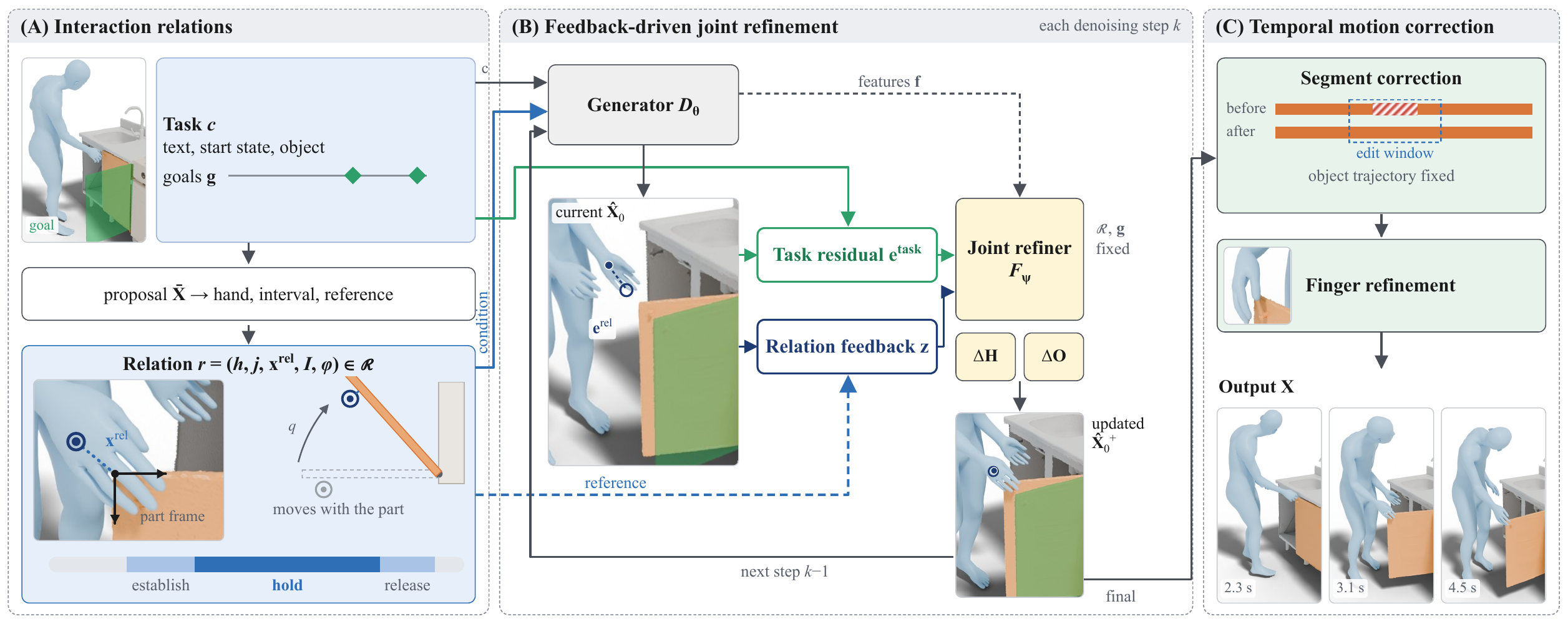}
  \fontsize{8}{8.4}\selectfont
  \captionsetup{skip=4pt,justification=raggedright,singlelinecheck=false}
\caption{\textbf{Overview of \ours.} \textbf{(A)} From the task conditions and a task-only proposal $\overline{\mathbf X}$, we construct relations $\Rset$: a hand, a wrist reference $\mathbf x^{\mathrm{rel}}$ in the part frame that moves with the part, and establish, hold, and release phases. \textbf{(B)} $\Rset$ conditions the generator $D_\theta$ and is the reference for relation feedback $\mathbf z$ (blue paths). At each denoising step, $F_\psi$ combines $\mathbf z$, task residuals $\mathbf e^{\mathrm{task}}$, and features $\mathbf f$ to update human and object motion, with $\Rset$ and goals $\mathbf g$ fixed. \textbf{(C)} Temporal correction then edits local hand motion with the object trajectory fixed followed by finger refinement.}
  \label{fig:architecture}
\end{figure*}

Given conditions $c=(\ell,\mathbf X_{\mathrm{init}},\mathbf g,\Mset)$ with a language instruction, an initial human--object state, sparse object targets, and an object model, \ours generates a coordinated motion sequence $\mathbf X=(\mathbf H,\mathbf O)$. The object goals specify object states at selected times; the intervening human and object trajectories are jointly generated and refined. To describe how the human participates in the process, we construct interaction relations $\Rset$ that specify which hand interacts with which object part, the wrist reference in the part frame, and the participation interval. During denoising, we measure deviations of the current motion prediction from the prescribed relations and object targets, and a learned refiner combines this feedback with motion context to predict joint human and object updates (\cref{fig:architecture}). After generation, local motion correction addresses remaining contact violations while keeping the resulting object trajectory and motion outside the editable region fixed. We describe relation-conditioned generation in \cref{sec:synthesis}, feedback-driven joint refinement in \cref{sec:refinement}, and constrained segment-level correction in \cref{sec:repair}.

\subsection{Preliminaries}
\label{sec:formulation}
\noindent\textbf{Representations.} We encode the language instruction $\ell$ with a frozen CLIP text encoder \cite{radford2021clip}, obtaining $\mathbf f^{\mathrm{text}}\in\mathbb R^{512}$. A length-$T$ human--object motion sequence is represented as $\mathbf X=(\mathbf H,\mathbf O)$, where $\mathbf H=\{\mathbf h_t\}_{t=1}^{T}$ and $\mathbf O=\{\mathbf o_t\}_{t=1}^{T}$. Each human state comprises joint positions $\mathbf P_t\in\mathbb R^{24\times3}$ and continuous 6D joint rotations $\boldsymbol\Theta_t\in\mathbb R^{22\times6}$ \cite{zhou2019continuity,li2024chois}. Flattening and concatenating these channels gives $\mathbf h_t\in\mathbb R^{204}$. For a free rigid object, the motion state is $\mathbf o_t=(\mathbf p_t,\mathbf R_t)$, with translation $\mathbf p_t\in\mathbb R^3$ and rotation $\mathbf R_t\in SO(3)$ encoded by its nine matrix entries. For a fixed-base articulated object, the motion state is $\mathbf o_t=q_t$, where $q_t\in\mathbb R$ specifies the angle or displacement of the active revolute or prismatic joint. We consider one active joint per target part. We distinguish the predicted position channels $\mathbf P_t$ from the decoded joint positions $\mathbf J_t=\FK_{\mathrm h} (\mathbf P_{t,\mathrm{root}},\boldsymbol\Theta_t)$. All geometric measurements of interaction realization and motion correction use this decoded geometry.

\noindent\textbf{Object geometry and kinematics.} The object model $\Mset$ provides the part geometry and, for articulated objects, the fixed base pose and joint kinematics. The pose of the interacting part $j$ in the common scene frame is
\begin{equation}
  (\mathbf p_{j,t},\mathbf R_{j,t})
  =\FK^{j}_{\Mset}(\mathbf o_t).
  \label{eq:fk}
\end{equation}
For rigid objects, this returns the represented object pose, while for articulated parts, it computes the pose from $q_t$. This part-pose interface allows relation measurements to be shared across rigid and articulated objects.

\noindent\textbf{Task conditions.} The task conditions provide the initial human--object state and sparse object targets $\mathbf g$ at specified frames. These targets constrain selected object-position coordinates and, for articulated parts, joint configurations. Their frame and coordinate masks identify the supplied values.

\subsection{Relation-Conditioned Motion Generation}
\label{sec:synthesis}
\label{sec:interaction}
To construct $\Rset$, we first sample a motion proposal $\overline{\mathbf X}$ using a motion generator conditioned only on $c$ without relation inputs. The proposal and task conditions are then used to construct the interaction relations. The relation-conditioned generator subsequently starts a separate sampling process from fresh noise under $c$ and $\Rset$.

\noindent\textbf{Interaction relations.} We represent each relation as $r=(h,j,\mathbf x^{\mathrm{rel}},I,\varphi)\in\Rset$, which specifies the participating hand, target part, part-local wrist reference, participation interval, and phase assignment. Proposal-derived relations describe the selected participation arrangement rather than contacts uniquely required by the task. The reference $\mathbf x_r^{\mathrm{rel}}\in\mathbb R^3$ specifies the wrist position relative to part $j$ during the hold phase. Hold denotes sustained hand--part participation, including pushing or pulling, without requiring a finger grasp. Expressing it in the part frame allows the reference to move with the interacting component, such as a translating drawer or a rotating door. This separates global part motion from deviations in relative hand following. It need not lie on the surface as the wrist is offset from the palm or finger contact. The time-dependent assignment $\varphi_r:I_r\rightarrow \{\mathrm{establish},\mathrm{hold},\mathrm{release}\}$ distinguishes approaching, maintaining, and leaving the relation, so that approaching and departing do not require the same fixed relative position as the hold phase. We denote each phase by $I_r^\nu=\{t\in I_r:\varphi_r(t)=\nu\}$. Each hand has at most one contiguous participation interval and one fixed local reference per task window, and an open end indicates that participation has not ended within the window.

The task specifies the target part, while wrist proximity in the motion proposal $\overline{\mathbf X}$ provides candidate hands and participation intervals. We compute each local reference as the coordinate-wise median of the proposal wrist positions, expressed in the target part frame over the candidate hold interval. Task-dependent coverage rules and initial geometry then fill in missing relations, extend participation to task-constrained frames, and replace missing or outlying references to create a base interval $I_r^{\mathrm{base}}$ and reference $\mathbf x_r^{\mathrm{rel}}$. A timing predictor $P_\eta$ trained separately on recorded participation intervals uses task conditions and initial geometry to conservatively extend the base intervals. This step preserves the selected hands and spatial references, after which the final phase bands are assigned deterministically. The resulting relations are encoded as per-frame participation, reference, and phase channels together with two-hand and foot-support flags. Hands, references, and intervals can be edited under the same task goals. During final sampling, the relation parameters and phase schedule remain fixed, while their realization is re-evaluated from the evolving motion prediction (\cref{sec:refinement}).

\noindent\textbf{Conditional generation.} We jointly generate human and object motion using a conditional diffusion model \cite{ho2020ddpm,tevet2023mdm,li2024chois}. For a normalized motion sequence $\mathbf X$, the forward process is $\mathbf X_k=\sqrt{\bar\alpha_k}\mathbf X +\sqrt{1-\bar\alpha_k}\boldsymbol\epsilon$, where $\boldsymbol\epsilon\sim\mathcal N(\mathbf 0,\mathbf I)$ and $\bar\alpha_k$ is the cumulative noise schedule. A temporal Transformer processes the human and object channels together over the full motion window, allowing their predictions to share temporal context. At each denoising step, the denoiser $D_\theta$ is conditioned on the task inputs $c$ and interaction relations $\Rset$. We also retain the per-frame hidden features for joint motion refinement:
\begin{equation}
  (\widehat{\mathbf X}_0,\mathbf f)
  =D_\theta(\mathbf X_k,k;c,\Rset),
  \label{eq:backbone}
\end{equation}
where $\mathbf f\in\mathbb R^{T\times d}$ denotes the motion features after geometric feature fusion and before the final output projection, and $d$ is the feature dimension. These features provide shape-aware motion context for the refiner, which additionally receives explicit measurements of how the current prediction realizes the prescribed relations and task targets (\cref{sec:refinement}). The text embedding provides sequence-level action context, while initial states, sparse targets, and interaction relations provide frame-level conditioning. To incorporate these conditions throughout the network, a Transformer control module processes the conditioned motion inputs in parallel with the motion Transformer, and injects its layer-wise features through zero-initialized linear projections \cite{zhang2023controlnet}. We also encode object shape using a basis point set representation \cite{prokudin2019bps} and fuse the resulting shape features with motion features through a geometry-aware module inspired by MaMi-HOI \cite{wang2026mamihoi}. Differences between learned motion and shape embeddings bias attention and contribute relative information to the residual feature update.

\subsection{Joint Human--Object Refinement}
\label{sec:refinement}
We refine human--object motion by jointly considering the selected interaction relations and object task targets. At each denoising step, we compute relation feedback and task residuals from the current clean-motion prediction $\widehat{\mathbf X}_0$. The refiner combines these signals with the generator's motion features to predict coordinated human and object updates across the motion sequence.

\noindent\textbf{Relation realization.} For relation $r$, let $\widehat{\mathbf w}_{h,t}$ be the FK-decoded wrist position and $(\widehat{\mathbf p}_{j,t},\widehat{\mathbf R}_{j,t})$ the decoded part pose. The local realization error is
\begin{equation}
  \mathbf e^{\mathrm{rel}}_{r,t}
  =\widehat{\mathbf R}_{j,t}^{\top}
   (\widehat{\mathbf w}_{h,t}-\widehat{\mathbf p}_{j,t})
   -\mathbf x_r^{\mathrm{rel}}.
  \label{eq:relgeom}
\end{equation}
This local error is invariant to a common rigid transformation of the wrist and part, separating global movement from relative following. Consistent relative following can coexist with a persistent surface gap, while surface proximity alone does not establish adherence to the selected relative motion. We therefore combine reference deviations with surface-proximity measurements. A deterministic extractor $\mathcal E_{\mathrm{rel}}$ forms per-frame, per-hand features $\mathbf z$ from the error, its magnitude and temporal change, its expression in the scene frame, the hand-to-surface distance, and the relation labels. Reference-based features are active only during participating hold frames, while surface distance remains observable
during approach and departure. These measurements capture proximity during approach and departure and relative following during hold.

\begin{table*}[!t]
  \centering
  \fontsize{8}{8.4}\selectfont
  \setlength{\tabcolsep}{2pt}
  \setlength{\aboverulesep}{1pt}
  \setlength{\belowrulesep}{1.5pt}
  \renewcommand{\arraystretch}{1.0}
  \captionsetup{skip=4pt,justification=raggedright,singlelinecheck=false}
  \caption{\textbf{Comparison on FullBodyManipulation dataset.} CoRelate achieves lower object-target errors and higher contact F1 and complete interaction success than the evaluated baselines. $\dagger$ denotes task-adapted methods.}
  \label{tab:omomo}
  \begin{tabular*}{\linewidth}{@{\extracolsep{\fill}}l*{15}{r}@{}}
    \toprule
    & \multicolumn{2}{c}{Object} & \multicolumn{4}{c}{Contact}
    & \multicolumn{5}{c}{Motion quality} & \multicolumn{3}{c}{GT fidelity} & Overall \\
    \cmidrule(lr){2-3}\cmidrule(lr){4-7}\cmidrule(lr){8-12}\cmidrule(lr){13-15}\cmidrule(l){16-16}
    Method & T\textsubscript{e}\,$\downarrow$ & T\textsubscript{xy}\,$\downarrow$ & C\textsubscript{prec}\,$\uparrow$ & C\textsubscript{rec}\,$\uparrow$ & C\textsubscript{F1}\,$\uparrow$ & MCC\,$\uparrow$
    & P\textsubscript{hand}\,$\downarrow$ & FS\,$\downarrow$ & FID\,$\downarrow$ & R\textsubscript{prec}\,$\uparrow$ & Div.\,$\rightarrow$
    & MPJPE\,$\downarrow$ & T\textsubscript{obj}\,$\downarrow$ & O\textsubscript{obj}\,$\downarrow$ & CIS\,$\uparrow$ \\
    \midrule
    MDM$^\dagger$ & 7.07 & 3.03 & 0.832 & 0.732 & 0.728 & 88.4 & 0.45 & 4.53 & 8.652 & 0.895 & 1.375 & 13.49 & 12.56 & 0.959 & 19.7 \\
    Pred-OMOMO & 7.55 & 3.48 & 0.825 & 0.696 & 0.710 & 83.2 & 0.39 & 4.25 & 3.621 & 0.858 & \textbf{1.353} & 14.38 & 13.17 & 0.964 & 9.9 \\
    CHOIS & 7.57 & 3.49 & 0.848 & 0.714 & 0.732 & 83.8 & 0.44 & \textbf{4.04} & 6.208 & 0.890 & 1.350 & 13.50 & 13.17 & 0.965 & 10.4 \\
    HOI-Dyn & 6.59 & 3.02 & 0.842 & 0.644 & 0.678 & 80.8 & \textbf{0.38} & 4.84 & 6.568 & 0.870 & 1.364 & 14.88 & 13.03 & 0.924 & 12.5 \\
    MaMi-HOI & 4.30 & 1.76 & \textbf{0.868} & 0.771 & 0.777 & 89.5 & 0.42 & 4.13 & 1.297 & 0.922 & 1.347 & \textbf{11.14} & 9.09 & \textbf{0.753} & 36.8 \\
    \midrule
    \textbf{\ours} & \textbf{0.85} & \textbf{0.49} & 0.866 & \textbf{0.885} & \textbf{0.854} & \textbf{95.3} & 0.42 & 4.06 & \textbf{0.921} & \textbf{0.945} & 1.344 & 11.32 & \textbf{8.22} & 0.760 & \textbf{71.2} \\
    \bottomrule
  \end{tabular*}
\end{table*}

\noindent\textbf{Task feedback and joint correction.} A relation discrepancy can be reduced by changing the human, the object, or both, but these adjustments can have different consequences for the supplied task targets. For example, moving an object toward the hand may close a contact gap while increasing its error at a specified waypoint. We therefore use task feedback to inform how human and object corrections are coordinated, measuring residuals at the supplied goal and waypoint coordinates, including articulated joint states:
\begin{equation}
  \begin{aligned}
    \mathbf z
      &=\mathcal E_{\mathrm{rel}}
        (\widehat{\mathbf X}_0,\Rset;\Mset),\\
    \mathbf e^{\mathrm{task}}
      &=\mathbf g-\mathcal P_{\mathrm{task}}
        (\widehat{\mathbf X}_0),\\
    (\Delta\mathbf H,\Delta\mathbf O)
      &=F_\psi(\mathbf f,\mathbf z,\mathbf e^{\mathrm{task}},k).
  \end{aligned}
  \label{eq:refine}
\end{equation}
The projection $\mathcal P_{\mathrm{task}}$ extracts the constrained coordinates at the specified frames using the same normalized units as $\mathbf g$. Constraint masks and time encodings accompany these residuals. Relation feedback describes how the selected participation arrangement is realized. The supplied targets describe the desired outcome, whereas their residuals describe the current prediction's deviation from it. A shared hand encoder and temporal attention fuse these inputs with the motion context $\mathbf f$. Two zero-initialized heads predict increments over the full motion window. The human head updates full-body joint-position and rotation channels, while the object head updates the object's native degrees of freedom. Both heads use shared relation, task, and temporal context to jointly predict the direction and magnitude of these increments. Temporal attention allows updates to account for the approach to contact, subsequent relative following, and transitions between them. We apply the predicted increments to obtain the corrected clean-motion prediction:
\begin{equation}
  \widehat{\mathbf X}_0^{+}
  =\operatorname{Update}
    (\widehat{\mathbf X}_0,\Delta\mathbf H,\Delta\mathbf O).
  \label{eq:motion-update}
\end{equation}
where $\operatorname{Update}$ applies increments according to the motion representation. Human increments act on joint-position and 6D rotation channels. Rigid-object updates modify translation and compose rotation with an incremental rotation parameterized in axis-angle form, while articulated updates modify the active joint state. The corrected prediction $\widehat{\mathbf X}_0^{+}$ is used in the diffusion sampling update. At the next denoising step, both feedback signals are recomputed from the new clean-motion prediction using the same relations and targets.

\noindent\textbf{Joint training.} We train the generator $D_\theta$ and joint refiner $F_\psi$ together on recorded motion sequences, with supervision applied through the corrected motion:
\begin{equation}
  \begin{aligned}
    \mathcal L_{\mathrm{gen}}
      ={}&\mathcal L_{\mathrm{motion}}
      +\lambda_{\mathrm{FK}}\mathcal L_{\mathrm{FK}}\\
      &+\lambda_{\mathrm{geom}}\mathcal L_{\mathrm{geom}}
      +\lambda_{\mathrm{sem}}\mathcal L_{\mathrm{sem}}.
  \end{aligned}
  \label{eq:gen-loss}
\end{equation}
These terms supervise motion reconstruction, FK-decoded joints, object geometry, and contact channels respectively. Supervision on the corrected motion trains the refiner to map relation deviations, task residuals, and motion context to coordinated human and object increments. Gradients jointly train $D_\theta$ and $F_\psi$, including through the geometric feedback. Relation construction lies outside this gradient path, and the timing predictor $P_\eta$ is trained separately. The separate proposal generator remains frozen. After motion generation, a finger-only optimization uses a training-derived PCA hand-pose prior and contact, penetration, release, and temporal-smoothness terms. It preserves body motion, wrist positions, palm orientations, and object trajectories.

\subsection{Temporal Motion Correction}
\label{sec:repair}
Joint refinement can leave local contact violations in the generated motion. Correcting these violations independently at each frame can introduce abrupt motion or discontinuities at edit boundaries. We therefore jointly optimize human joint rotations over short temporal segments around contact violations while keeping the generated object trajectory fixed, preserving its existing object-target errors.

\noindent\textbf{Editable regions.} Given a post-refinement prediction $\mathbf X_{\mathrm{ref}}$, the selected relations identify the participating hands and intervals, while the current geometry locates contact violations during object motion. An editable mask $M_{\mathrm{edit}}$ specifies which body joints and frames can change within these temporal segments. We extend each interval to include approach and boundary-transition frames. The observed prefix and motion outside the editable region remain protected.

\noindent\textbf{Temporal optimization.} At inference time, the solver jointly optimizes shoulder, elbow, and wrist rotations over each selected segment, keeping network parameters, root translation, and object motion fixed. FK updates hand positions while preserving bone lengths. Its objective combines geometric, temporal, and edit-magnitude terms:
\begin{equation}
  \begin{aligned}
    \mathcal L_{\mathrm{seg}}={}&w_c\mathcal L_c+w_s\mathcal L_s+w_p\mathcal L_p\\
      &+w_a\mathcal L_a+w_j\mathcal L_j+w_d\mathcal L_d.
  \end{aligned}
  \label{eq:segment-loss}
\end{equation}
The contact and slip terms $\mathcal L_c$ and $\mathcal L_s$ encourage surface proximity and relative following. The temporal terms $\mathcal L_a$ and $\mathcal L_j$ penalize changes in acceleration and excessive wrist steps, including transitions to unchanged motion. A surface-proximity proxy $\mathcal L_p$ and an edit-magnitude penalty $\mathcal L_d$ regularize the correction. Joint optimization distributes adjustments across neighboring frames. The correction preserves object trajectories and protected human motion, leaving object-target errors unchanged.
\section{Experiments}
\label{sec:experiments}

\begin{table*}[!t]
  \centering
  \fontsize{8}{8.4}\selectfont
  \setlength{\tabcolsep}{2pt}
  \setlength{\aboverulesep}{1pt}
  \setlength{\belowrulesep}{1.5pt}
  \renewcommand{\arraystretch}{1.0}
  \captionsetup{skip=4pt,justification=raggedright,singlelinecheck=false}
  \caption{\textbf{Comparison on ParaHome dataset.} CoRelate achieves higher task completion, contact F1, and complete interaction success than the evaluated baselines. $\dagger$ denotes task-adapted methods.}
  \label{tab:parahome}
  \begin{tabular*}{\linewidth}{@{\extracolsep{\fill}}l*{12}{r}@{}}
    \toprule
    & Task & \multicolumn{4}{c}{Contact} & \multicolumn{6}{c}{Motion quality} & Overall \\
    \cmidrule(lr){2-2}\cmidrule(lr){3-6}\cmidrule(lr){7-12}\cmidrule(l){13-13}
    Method & completion\,$\uparrow$ & C\textsubscript{acc}\textsuperscript{tem}\,$\uparrow$ & C\textsubscript{acc}\textsuperscript{body}\,$\uparrow$ & C\textsubscript{F1}\,$\uparrow$ & MCC\,$\uparrow$
    & Pene\,$\downarrow$ & FS\,$\downarrow$ & Jerk\textsubscript{obj}\,$\downarrow$ & FID\,$\downarrow$ & R\textsubscript{prec}\,$\uparrow$ & Div.\,$\rightarrow$ & CIS\,$\uparrow$ \\
    \midrule
    CoDA$^\dagger$ & 22.0 & 0.502 & 0.918 & 0.407 & 56.7 & 0.872 & \textbf{0.180} & \textbf{0.488} & 0.227 & 0.560 & 1.197 & 13.2 \\
    SK-HOI$^\dagger$ & 33.0 & 0.425 & \textbf{0.920} & 0.425 & 35.0 & 0.743 & 0.565 & 0.589 & 0.385 & 0.340 & 1.159 & 16.2 \\
    \midrule
    \textbf{\ours} & \textbf{92.0} & \textbf{0.652} & 0.910 & \textbf{0.775} & \textbf{71.7} & \textbf{0.721} & 0.225 & 0.546 & \textbf{0.117} & \textbf{0.840} & \textbf{1.271} & \textbf{36.0} \\
    \bottomrule
  \end{tabular*}
\end{table*}

\begin{figure*}[!t]
  \centering
  \includegraphics[width=\linewidth]{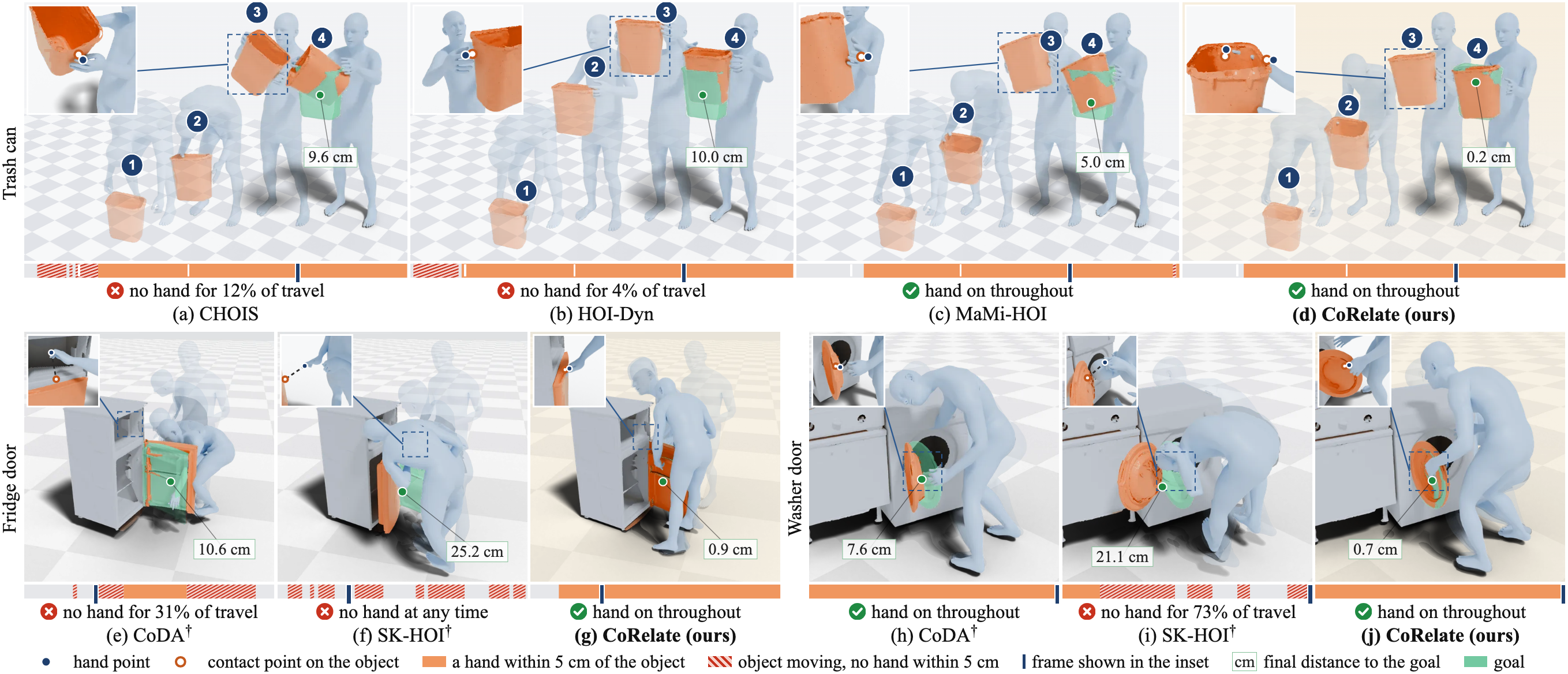}
  \fontsize{8}{8.4}\selectfont
  \captionsetup{skip=4pt,justification=raggedright,singlelinecheck=false}
  \caption{\textbf{Qualitative comparisons.} We mark the final distance to the goal and the frames where the object moves without a hand. \ours maintains hand contact as objects move toward their goals.}
  \label{fig:qualitative}
  \vspace{-10pt}
\end{figure*}

\subsection{Experimental Setup}
\label{sec:setup}

\noindent\textbf{Datasets.}
We evaluate on FullBodyManipulation \cite{li2023omomo} and ParaHome \cite{kim2025parahome}. FullBodyManipulation contains approximately 10 hours of paired human and object motion involving 15 objects. We follow the CHOIS subject split \cite{li2024chois}. ParaHome captures household interactions involving rigid objects and articulated parts. We use recording-disjoint splits for evaluation. Models are trained independently on each dataset.

\noindent\textbf{Metrics.}
We measure object-target accuracy using endpoint position error T\textsubscript{e} and horizontal waypoint error T\textsubscript{xy}, following prior protocols \cite{li2024chois,wang2026mamihoi}. Interaction is evaluated using hand-contact precision, recall, and F1 (C\textsubscript{prec}, C\textsubscript{rec}, C\textsubscript{F1}) against reference contact labels, averaged over windows. We propose Motion-weighted Contact Coverage (MCC), which measures the fraction of object motion accompanied by hand contact. We assess geometry and motion quality using hand-mesh penetration (P\textsubscript{hand}), foot sliding (FS), FID, top-3 text--motion retrieval (R\textsubscript{prec}), and embedding-space diversity (Div.). Reference fidelity is measured by root-relative MPJPE, object translation error T\textsubscript{obj}, and rotation-matrix error O\textsubscript{obj}. Position and penetration errors are in cm, and O\textsubscript{obj} is dimensionless. We also define Complete Interaction Success (CIS), which measures the percentage of outputs that jointly satisfy object goals and waypoints, applicable contact requirements, and motion-validity criteria. For ParaHome, we additionally include temporal and body contact accuracy (C\textsubscript{acc}\textsuperscript{tem}, C\textsubscript{acc}\textsuperscript{body}), marker penetration (Pene), and object-keypoint jerk (Jerk\textsubscript{obj})~\cite{peng2026skhoi}. Task completion measures whether rigid objects reach their target positions or articulated parts reach their target joint states.

\noindent\textbf{Baselines.}
On FullBodyManipulation, we compare with MDM \cite{tevet2023mdm}, Pred-OMOMO \cite{li2023omomo,li2024chois}, CHOIS \cite{li2024chois}, MaMi-HOI \cite{wang2026mamihoi}, and HOI-Dyn \cite{wu2025hoidyn}. MDM is adapted to jointly generate human and object motion conditioned on the object goals and waypoints. Pred-OMOMO generates human motion conditioned on an object trajectory predicted by CHOIS. On ParaHome, we compare with CoDA \cite{pi2025coda} and SK-HOI \cite{peng2026skhoi}. CoDA is adapted with object-goal conditioning. For SK-HOI, since the released codebase lacks data preprocessing and contact-field construction, we reimplement these components and adapt the released cascade to use predicted object trajectories and contacts at inference. These baselines cover object-motion control, contact-aware generation, and human--object consistency, which provide complementary comparisons for target-goal satisfaction and human participation.

\noindent\textbf{Implementation.}
The generator uses four Transformer layers with hidden dimension 512, while the refiner uses one temporal layer with hidden dimension 256. Both employ four attention heads. We use 120-frame windows at 30 fps and 1,000 diffusion steps per sampling pass. All experiments are conducted on a single NVIDIA RTX 4090 GPU.

\begin{table*}[!t]
  \centering
  \fontsize{8}{8.4}\selectfont
  \setlength{\tabcolsep}{2pt}
  \setlength{\aboverulesep}{1pt}
  \setlength{\belowrulesep}{1.5pt}
  \renewcommand{\arraystretch}{1.05}
  \captionsetup{skip=4pt,justification=raggedright,singlelinecheck=false}
  \caption{\textbf{Ablation study.} Each design choice contributes and task and relation feedback show complementary gains when combined.}
  \label{tab:ablation}
  \begin{tabular*}{\linewidth}{@{\extracolsep{\fill}}l*{12}{r}@{}}
    \toprule
    & \multicolumn{2}{c}{Object} & \multicolumn{2}{c}{Contact}
    & \multicolumn{4}{c}{Motion quality} & \multicolumn{3}{c}{GT fidelity} & Overall \\
    \cmidrule(lr){2-3}\cmidrule(lr){4-5}\cmidrule(lr){6-9}\cmidrule(lr){10-12}\cmidrule(l){13-13}
    Configuration & T\textsubscript{e}\,$\downarrow$ & T\textsubscript{xy}\,$\downarrow$
    & C\textsubscript{F1}\,$\uparrow$ & MCC\,$\uparrow$
    & P\textsubscript{hand}\,$\downarrow$ & FS\,$\downarrow$
    & FID\,$\downarrow$ & R\textsubscript{prec}\,$\uparrow$
    & MPJPE\,$\downarrow$ & T\textsubscript{obj}\,$\downarrow$ & O\textsubscript{obj}\,$\downarrow$ & CIS\,$\uparrow$ \\
    \midrule
    Base & 3.17 & 1.83 & 0.761 & 89.4 & 0.44 & 4.15 & 1.248 & 0.939 & 12.01 & 8.78 & 0.748 & 41.2 \\
    + Relation conditioning & 2.93 & 1.74 & 0.821 & 94.3 & \textbf{0.41} & \textbf{3.97} & \textbf{0.902} & 0.934 & 11.51 & 8.91 & 0.754 & 46.4 \\
    + Joint refiner w/o feedback & 2.97 & 1.74 & 0.821 & 93.4 & 0.43 & 4.02 & 0.926 & 0.927 & 11.47 & 8.85 & \textbf{0.751} & 47.8 \\
    + Task feedback & 0.93 & 0.75 & 0.825 & 94.2 & \textbf{0.41} & 4.13 & 1.116 & 0.932 & 11.50 & 8.58 & 0.753 & 66.4 \\
    \textbf{+ Relation feedback (Full)} & \textbf{0.85} & \textbf{0.49} & \textbf{0.854} & \textbf{95.3} & 0.42 & 4.06 & 0.921 & \textbf{0.945} & \textbf{11.32} & \textbf{8.22} & 0.760 & \textbf{71.2} \\
    \bottomrule
  \end{tabular*}
\end{table*}

\subsection{Quantitative Results}
\label{sec:quant}
\Cref{tab:omomo} compares human--object interaction synthesis under sparse object goals. \ours achieves the lowest endpoint and waypoint errors and the highest contact F1 and motion-weighted contact coverage. These improvements indicate more accurate object control and stronger hand--object engagement during object motion. The highest CIS further shows that \ours more consistently satisfies task, contact, and motion-validity requirements jointly. \ours also achieves the highest text--motion retrieval accuracy, which indicates strong alignment between the generated interactions and the input descriptions. FID, foot sliding, and hand penetration remain competitive alongside these improvements. The lowest object translation error further shows that more accurate goal satisfaction is accompanied by closer agreement with the reference object trajectories.

\Cref{tab:parahome} evaluates rigid and articulated household interactions, where human participation must be coordinated with object displacement or changes in articulation state. \ours achieves higher task completion, temporal contact accuracy, contact F1, and motion-weighted contact coverage than the task-adapted CoDA and SK-HOI baselines. These results support improved object-task execution accompanied by more consistent human contact during object motion. The gains do not extend to every individual measure; body contact accuracy, foot sliding, and object jerk favor different baselines. \ours also obtains lower FID and higher text--motion retrieval accuracy. Its higher CIS further indicates that task, contact, and motion-validity requirements are better satisfied jointly.

\vspace{-3pt}
\subsection{Qualitative Results}
\label{sec:qual}
We visualize the generated human--object interactions in \cref{fig:qualitative}. In the rigid-object example, CHOIS and HOI-Dyn exhibit intervals of object motion without hand contact and leave the object away from its goal. MaMi-HOI maintains hand contact but still leaves a visible offset between the final object position and the goals. Similar coordination issues appear in the articulated interactions. CoDA makes contact with the manipulated doors, but contact can be interrupted and the doors do not reach their target configurations. SK-HOI shows more pronounced hand--part separation, with the door moving despite absent or lost hand contact. In comparison, \ours maintains hand contact as the object or articulated part moves toward its target and produces interactions where human participation accompanies task completion.

\subsection{Ablation Study}
\label{sec:ablation}
We assess the cumulative contributions of generation and refinement components on FullBodyManipulation (\cref{tab:ablation}). Relation conditioning improves contact metrics and complete interaction success, which supports the use of explicit participation information to guide generation. Adding the refiner without feedback provides only a modest additional success gain, whereas task feedback provides the largest improvement in complete interaction success and substantially reduces object-goal errors. This contrast highlights the value of explicit task residuals beyond adding the refinement architecture alone. Relation feedback further improves both contact metrics and object-goal accuracy, which suggests that measuring interaction realization helps refine contact together with the intended object motion. The full model achieves the highest complete interaction success. R-precision remains relatively stable, and relation feedback improves FID over task feedback alone, although motion-quality metrics do not improve uniformly across component additions. These results support relation conditioning and complementary task and relation feedback.

\begin{figure}[!htbp]
  \centering
  \includegraphics[width=\linewidth]{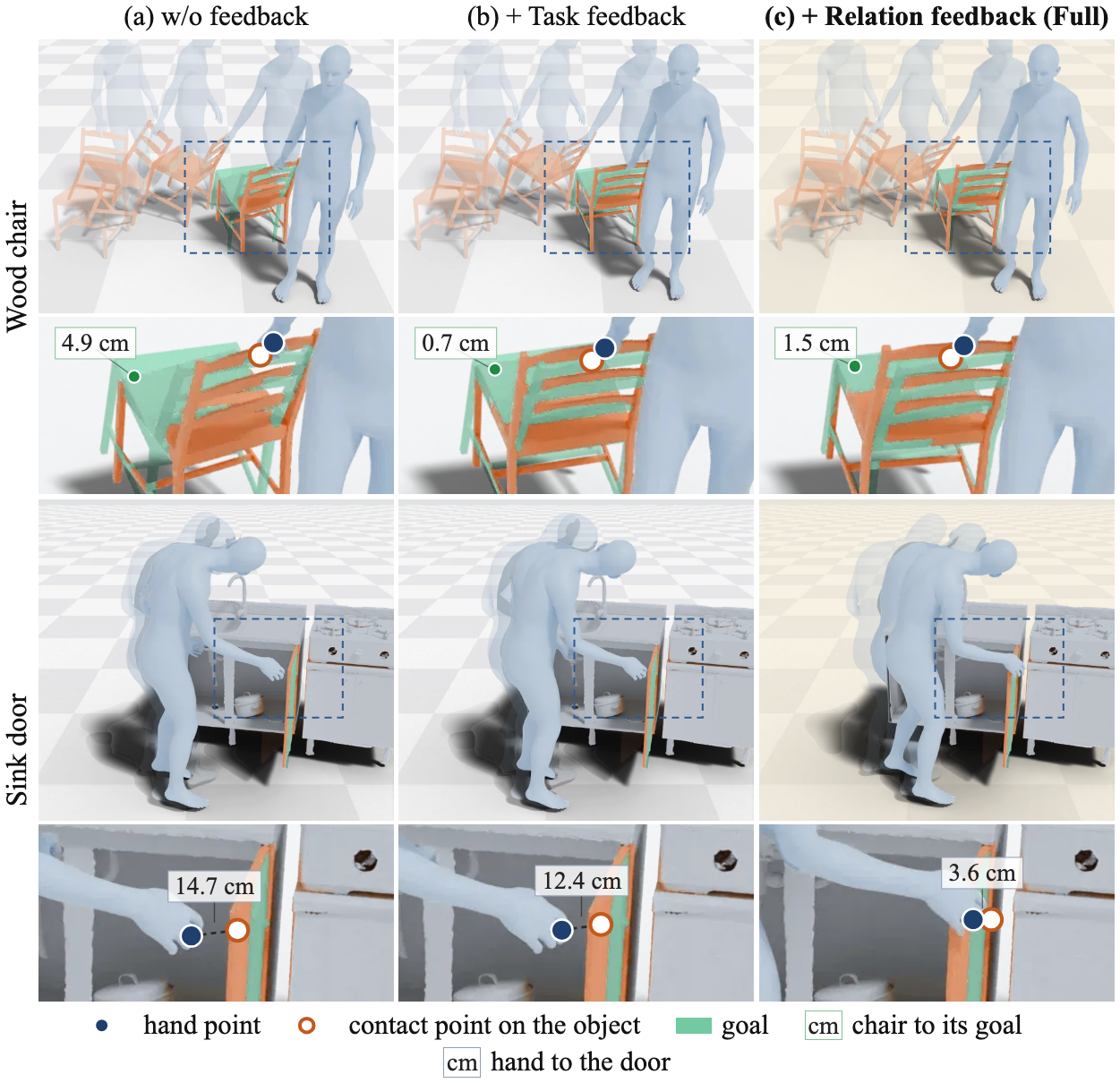}
  \fontsize{8}{8.4}\selectfont
  \captionsetup{skip=4pt,justification=raggedright,singlelinecheck=false}
  \caption{\textbf{Qualitative ablation visualization.} We progressively add task and relation feedback to the no-feedback baseline and highlight incorrect interactions.}
  \label{fig:ablation}
\end{figure}

\section{Conclusion and Limitations}
\label{sec:conclusion}

We presented \ours for coordinated human--object motion synthesis under sparse object goals. Explicit interaction relations serve as both generation conditions and references for measuring interaction realization. Relation feedback and task residuals jointly guide human--object updates during denoising, while temporal motion correction addresses remaining contact failures without changing object trajectories or protected human motion.

Our method has several limitations. Relations remain fixed during generation, which may limit recovery from unsuitable participation arrangements, and the current representation does not support repeated regrasping within a motion window. Local motion correction cannot resolve failures that require changes to object motion or broader body adjustments. The method also does not explicitly model articulated finger grasps or contact forces, so geometric consistency alone does not guarantee physical validity. Future work could adapt participation timing and hand--part assignments during generation to support repeated regrasping.
\FloatBarrier

{
    \small
    \bibliographystyle{ieeenat_fullname}
    \bibliography{main}
}

\end{document}